\documentclass[10pt,twocolumn,letterpaper]{article}

\usepackage{cvpr}
\usepackage{times}
\usepackage{epsfig}
\usepackage{graphicx}
\usepackage{amsmath}
\usepackage{amssymb}

\usepackage{booktabs}
\usepackage{gensymb}
\usepackage{multirow}
\usepackage{authblk}
\usepackage{url}
\usepackage[T1]{fontenc}
\usepackage{pifont}
\newcommand{\cmark}{\ding{51}}
\newcommand{\xmark}{\ding{55}}
\usepackage{titling}
\def\eg{e.g.,\ }

\usepackage[pagebackref=true,breaklinks=true,letterpaper=true,colorlinks,bookmarks=false]{hyperref}

\makeatletter
\renewcommand\AB@affilsepx{, \protect\Affilfont}
\makeatother

\cvprfinalcopy 

\def\cvprPaperID{} 

\ifcvprfinal\fi
\begin{document}

\title{From Image Collections to Point Clouds \\ with Self-supervised Shape and Pose Networks}

\author{\vspace{-1em}%
K L Navaneet$^1$\quad 
Ansu Mathew$^1$\quad 
Shashank Kashyap$^1$\quad 
Wei-Chih Hung$^2$\\ \vspace{-0.0em}%
Varun Jampani$^3$\quad 
R. Venkatesh Babu$^1$\\ \vspace{1em}%
$^1$Indian Institute of Science \qquad
$^2$University of California, Merced \qquad 
$^3$Google Research\\
}

\predate{}
\date{}
\postdate{}
\maketitle
\thispagestyle{empty}

\begin{abstract}
  Reconstructing 3D models from 2D images is one of the fundamental problems in computer vision. In this work, we propose a deep learning technique for 3D object reconstruction from a single image. Contrary to recent works that either use 3D supervision or multi-view supervision, we use only single view images with no pose information during training as well. This makes our approach more practical requiring only an image collection of an object category and the corresponding silhouettes. We learn both 3D point cloud reconstruction and pose estimation networks in a self-supervised manner, making use of differentiable point cloud renderer to train with 2D supervision. A key novelty of the proposed technique is to impose 3D geometric reasoning into predicted 3D point clouds by rotating them with randomly sampled poses and then enforcing cycle consistency on both 3D reconstructions and poses. In addition, using single-view supervision allows us to do test-time optimization on a given test image. Experiments on the synthetic ShapeNet and real-world Pix3D datasets demonstrate that our approach, despite using less supervision, can achieve competitive performance compared to pose-supervised and multi-view supervised approaches.
\end{abstract}

\section{Introduction}

3D object reconstruction is a long standing problem in the field of computer vision. With the success of deep learning based approaches, the task of single image based 3D object reconstruction has received significant attention in the recent years. The problem has several applications such as view synthesis and  
grasping and manipulation of objects. 

Early works~\cite{girdhar2016learning,choy20163d,fan2017point} on single image based 3D reconstruction utilize full 3D supervision in the form of 3D voxels, meshes or point clouds. However, such approaches require large amounts of 3D data for training, which is hard and expensive to obtain. Several recent works~\cite{yan2016perspective,tulsiani2017multi} have focused on utilizing multi-view 2D supervision in the form of color images and object silhouettes as an effective alternative training protocol. 
A key component in these techniques is the differentiable rendering module that enables the use of 2D observations as supervision using reprojection consistency based losses.~
However, most of these approaches require multiple 2D view of the same 3D model along with the associated camera pose information in the training stage.
This is a major limitation in applying these techniques in a practical setting where such supervisory data is difficult to obtain.      

\begin{figure}[t]
    \centering
    \includegraphics[width=1.0\linewidth]{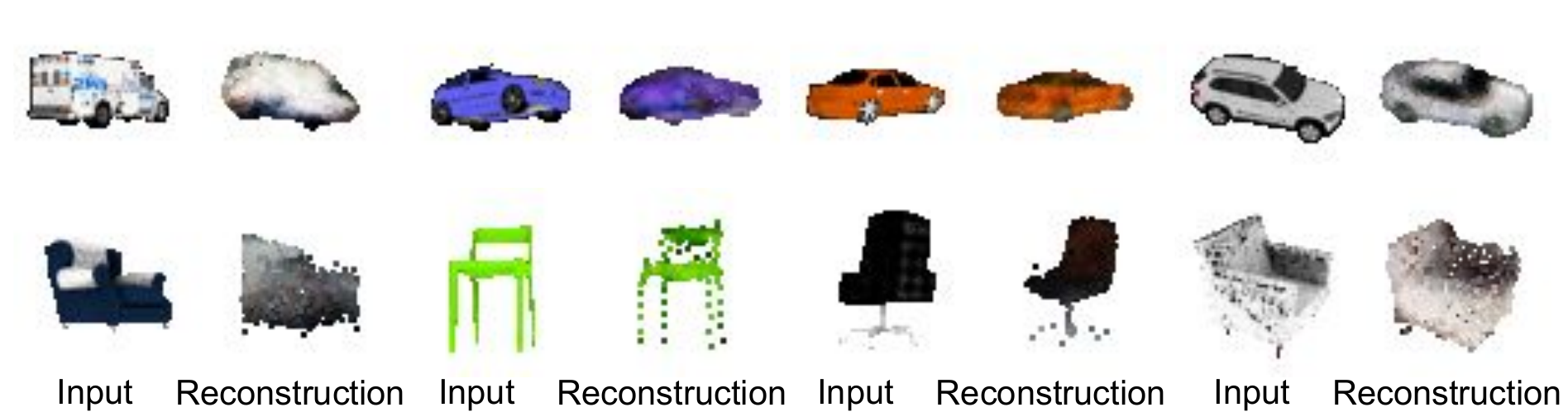}
    \caption{\textbf{Single-image 3D Reconstructions.} Input image and corresponding projection from reconstructed 3D point clouds. We reconstruct the 3D output from a single input image using a completely self-supervised approach.}
    \label{fig:teaser}
\end{figure}
In this work, we set out to tackle a more challenging problem of learning 3D object reconstructions from image and corresponding silhouette collections. Given a collection of images and corresponding object silhouettes belonging to the same object category such as car, with just a single view from each object instance and no ground truth camera pose information, our goal is to learn 3D object reconstructions (Fig.~\ref{fig:teaser}). The proposed approach is practically useful and enables us to make effective use of the large amounts of 2D training data for learning 3D reconstructions. Since it is possible to easily obtain object silhouettes in the absence of ground truth masks, here we make the reasonable assumption that the image collection contains corresponding silhouettes.
A key challenge in our training setting is to simultaneously learn both camera pose estimation and 3D reconstruction while avoiding degenerate solutions. For instance, a degenerate solution for 3D reconstruction would be to just lift 2D pixels in a given image onto a 3D plane. Although such a flat 3D reconstruction perfectly explains a given image, that is obviously not a desired 3D shape. 
In this work, we introduce loss functions that are tailored towards simultaneous learning of the pose and reconstruction networks while avoiding such degenerate solutions. Specifically, we propose to use
geometric and pose cycle consistency losses.
To enforce geometric cycle consistency, we make use of the fact that multiple 2D views from the same 3D model must all result in the same 3D model upon reconstruction. However, note that these multiple 2D views are intermediate representations obtained in our framework utilizing just a single image per model. To correctly regress the pose values, we 
obtain projections from random viewpoints to enforce consistency in pose predictions. Motivated by the observation that the reconstruction performance improves remarkably when multiple 2D views are used for supervision, we aim to utilize additional images as supervision. However, since our problem setting limits the number of views from each 3D model to one, we effectively retrieve images from the training set with similar 3D geometry in a self-supervised manner. We utilize them as additional supervision in the form of cross-silhouette consistency to aid the training of pose and reconstruction networks. 

Since our approach is self-supervised, we can adapt our network to obtain better reconstructions on a given test input image by performing additional optimization during inference.
We propose regularization losses to avoid over-fitting on a single test sample. This ensures that the 3D reconstructions are more accurate from input viewpoint while maintaining their 3D structure in the occluded regions. 

We benchmark our approach on the synthetic ShapeNet~\cite{chang2015shapenet} dataset and observe that it achieves comparable performance to the state-of-the-art multi-view supervised approaches~\cite{navaneet2019differ,insafutdinov2018pointclouds}. We also evaluate our approach on the real-world Pix3D~\cite{pix3d} dataset and show comparable or improved performance over a pose supervised approach~\cite{navaneet2019differ}. We also present possible applications of our approach for dense point correspondence and 3D semantic part-segmentation. To the best of our knowledge, this is the first completely self-supervised approach for 3D point cloud reconstruction from image and silhouette collections.

To summarize, we make the following contributions in this work: 
\begin{itemize}
\itemsep-0.1em
    \item We propose a framework to achieve single image 3D point cloud reconstruction in a completely self-supervised manner. 
    \item We introduce cycle consistency losses on both pose and 3D reconstructions 
    to aid the training of the pose and reconstruction networks respectively.
    \item We effectively mine images from geometrically similar models to enforce cross-silhouette consistency, leading to significantly improved reconstructions 
    \item We perform thorough evaluations to demonstrate the efficacy of each component of the proposed approach on the ShapeNet dataset. We also achieve competitive performance to pose and multi-view supervised approaches on both ShapeNet and real-world Pix3D datasets.
\end{itemize}

\begin{figure*}[t]
    \centering
        \includegraphics[width=1.0\linewidth]{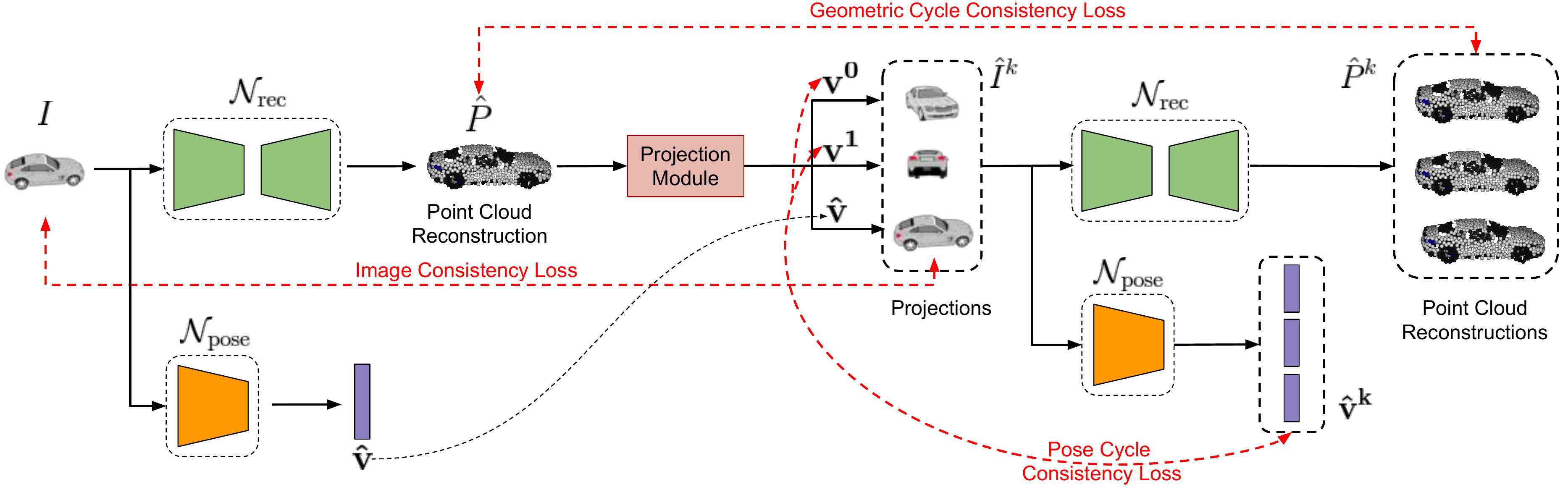}
    \caption{\textbf{Approach Overview.} We propose a cycle consistency based approach to obtain 3D reconstructions from a collection of images and their corresponding foreground masks. An encoder-decoder architecture based network is used to regress the 3D coordinates of the point cloud reconstruction $\hat{P}$. A pose network is used to obtain 3D camera pose predictions $\mathbf{\hat{v}}$ from the input image. DIFFER~\cite{navaneet2019differ} is used to render the reconstruction in the predicted viewpoint. Additionally, reconstructions are also projected from randomly sampled poses to obtain $k$ projections which are again used to reconstruct $k$ point clouds $\hat{P}^k$. We enforce a 3D cycle consistency loss on $\hat{P}$ and $\hat{P}^k$ to train $\mathcal{N}_{\text{rec}}$. Similarly the randomly sampled poses and corresponding projections are considered as pseudo ground truth labels to enforce pose cycle consistency loss. The red dashed arrows in the diagram indicate the proposed losses.} 
    \label{fig:net_arch}
\end{figure*}
\section{Related Works}
\vspace{-1mm}
\noindent \textbf{Single Image Based 3D Reconstruction}
Several learning based works in the recent past have tackled the problem of single image based 3D object reconstruction. The initial works~\cite{girdhar2016learning,choy20163d,fan2017point,tatarchenko2017octree,groueix2018,mandikal20183dlmnet} make use of full 3D supervision in terms of ground-truth voxels or point clouds. 
Choy \etal~\cite{choy20163d} utilize multiple inputs for improved voxel reconstructions. Fan \etal~\cite{fan2017point} is one of the first works to learn point cloud reconstructions from images using a deeply learned network. They made use of set distance based losses to directly regress the 3D locations of the points. 
Mandikal \etal~\cite{mandikal20183dpsrnet} extend ~\cite{fan2017point} to predict point clouds with part segmentations using a part-aware distance metric calculation. \\

\vspace{-2mm}
\noindent \textbf{2D Supervised Approaches}
While the above works obtain promising results, they require ground truth 3D models as supervision which is complex and expensive to obtain. To overcome this, several works~\cite{yan2016perspective,tulsiani2017multi,wu2017marrnet,zhu2017rethinking,lin2018learning,navaneet2019capnet,henderson2018learning,tulsiani2018unsup,insafutdinov2018pointclouds,navaneet2019differ,kato2018renderer} have explored 2D supervised approaches utilizing 2D images, silhouettes, depth maps and surface normal maps. These works aim to develop ways to go from the 3D representation to the 2D projections in a differentiable manner in order to effectively back-propagate the gradients from the 2D loss functions to the reconstruction network. Yan \etal~\cite{yan2016perspective} achieve this on voxel based 3D representations by performing grid sampling of voxels to obtain foreground mask projections. Re-projection losses are used from multiple viewpoints to train the network. Similarly, Tulsiani \etal~\cite{tulsiani2017multi} use a differentiable ray consistency based loss to reconstruct not only the shape information, but also features like color. The work is extended in ~\cite{tulsiani2018unsup} where a multiple-view consistency based loss is formulated to simultaneously predict 3D camera pose and object reconstructions. 
Motivated by the computational and performance advantages offered by point clouds, a number of works have sought to design rendering modules for projecting 3D points. Insafutdinov and Dosovitskiy~\cite{insafutdinov2018pointclouds} and Navaneet \etal~\cite{navaneet2019capnet,navaneet2019differ} develop differentiable projection modules to project points and corresponding features on to the 2D plane, enabling training on 2D representations like silhouettes, depth maps, images and part segmentations.

\noindent \textbf{Weakly Supervised Approaches}
Among the weakly supervised approaches, ~\cite{kanazawa2018learning,mees2019self,Li_2019_CVPR,navaneet2019differ,tulsiani2018unsup,insafutdinov2018pointclouds} are the closest to ours. 
Mees \etal~\cite{mees2019self} utilize mean 3D object models to learn 3D reconstructions in a self-supervised manner. Li \etal~\cite{Li_2019_CVPR} generate 3D models using a self-supervised approach, but do not perform reconstruction from RGB images. In SSL-Net~\cite{sun2019ssl}, 3D models are used to pre-train one of the networks before performing self-supervised reconstruction. To the best of our knowledge, we are the first to obtain colored 3D point cloud reconstructions from just a collection of images and corresponding silhouettes. 

\section{Approach}
\label{sec:approach}
We aim to obtain 3D point cloud reconstruction from a single image in a self-supervised setting. To this end, we propose a learning based approach with an encoder-decoder architecture based network to predict the reconstructions. Let $I$ be the image input to the network, $M$ the foreground object mask and $\hat{P} \in \mathbb{R}^{N\times3}$ the corresponding point cloud reconstruction obtained using the reconstruction network $\mathcal{N}_{\text{rec}}$  (refer Fig.~\ref{fig:net_arch}). $N$ is the number of points in the reconstructed point cloud. In the absence of ground truth 3D models, all our supervisory data, which is the set of input images and corresponding silhouettes, lies in the 2D domain. In order to utilize these 2D observations to train the network, we would need to project the reconstructed point cloud on to the 2D image plane. We use the differentiable projection modules proposed in DIFFER~\cite{navaneet2019differ} and CAPNet~\cite{navaneet2019capnet} to obtain color and mask projections respectively from a given viewpoint. The viewpoint $\mathbf{v}$ associated with the input image is characterized by azimuth and elevation values of 
the camera in the 3D space placed at a fixed distance from the object. We use another encoder network $\mathcal{N}_{\text{pose}}$ to obtain the viewpoint prediction $\mathbf{\hat{v}}$. 
The reconstructed point cloud is projected from the predicted viewpoint using the differentiable projection module to obtain 2D image and mask predictions $\hat{I}$ and $\hat{M}$ respectively. If the predicted viewpoint and reconstructions are correct, the 2D projections will match the input image. To enforce this, we use the losses proposed in DIFFER~\cite{navaneet2019differ} to optimize both the reconstruction and pose prediction networks. Specifically, we use the following image ($\mathcal{L}_\text{I}$) and mask ($\mathcal{L}_\text{M}$) loss functions: 
\begin{equation}
    \mathcal{L}_{\text{I}} = \frac{1}{hw}\sum\limits_{i,j} ||I_{i,j}-\hat{I}_{i,j}||^{2}_{2}
    \label{eq:rgb_loss}
\end{equation}

\begin{align}
    \mathcal{L}_{\text{bce}} &= \frac{1}{hw}\sum\limits_{i,j} -M_{i,j} \text{log}\hat{M}_{i,j} - (1-M_{i,j}) \text{log}(1-\hat{M}_{i,j}) \label{eq:bce_loss_0} \\
    \mathcal{L}_{\mathrm{aff}} &= \sum\limits_{i,j} \min\limits_{(k,l)\in M_{+}} ((i-k)^2 + (j-l)^2)\hat{M}_{i,j} M_{k,l} \nonumber \\ 
    & + \sum\limits_{i,j} \min\limits_{(k,l)\in \hat{M}_{+}} ((i-k)^2 + (j-l)^2)M_{i,j}\hat{M}_{k,l}
    \label{eq:loss_aff} \\
    \mathcal{L}_\text{M} &= \mathcal{L}_{\text{bce}} + \mathcal{L}_{\text{aff}} \label{eq:bce_loss}
\end{align}

where $h,w$ are the height and width of the 2D observations respectively. $M_+$ and $\hat{M}_+$ are sets of pixel coordinates of the ground truth and predicted projections whose values are non-zero. In this formulation, the predictions by the reconstruction and pose networks rely heavily on each other. Since the predicted viewpoint is used in projection, the reconstruction network can predict correct 3D models that consistently match the input image only if the pose predictions are accurate. Similarly, since the pose network parameters are optimized using projection losses, the predicted pose values will be correct only if the reconstructions are reasonable. In such a situation, the network can collapse to degenerate solutions. 
For instance, the predicted viewpoint can be constant regardless of the input and the 3D reconstruction can be planar. The networks would still achieve zero loss as long as they reproduce the input image from the predicted constant viewpoint. To avoid such degenerate solutions, we propose novel cycle consistency losses to train both reconstruction and pose networks.

\subsection{Geometric Cycle Consistency Loss}
\label{subsec:3d_cyclic_consistency}
We propose geometric cycle consistency loss to train the reconstruction network (Fig.~\ref{fig:net_arch}) to avoid degenerate reconstructions. 
The reconstructed point cloud $\hat{P}$ is projected from $k$ randomly sampled viewpoints $\{\mathbf{v}^{i}\}_{1}^{k}$. Let $\{\hat{I}^{i}\}_{1}^{k}$ be the corresponding image projections. These images are used as input to the reconstruction network $\mathcal{N}_{\text{rec}}$ and the corresponding reconstructed point clouds $\{\hat{P}^{i}\}_{1}^{k}$ are obtained. Since each of the projections and the input image are all associated with the same 3D object, the corresponding point clouds must all be consistent with each other. To enforce this, we define the geometric cycle consistency loss as follows:
\begin{equation}
    \mathcal{L}_{\text{G}} = \sum \limits_{i=1}^{k}\text{d}_\text{Ch}(\hat{P}, \hat{P}^{i})
\end{equation}
where $\text{d}_\text{Ch}(\cdot,\cdot)$ denotes the Chamfer distance between two point clouds. The reconstruction network is trained using a combination of the mask and image losses and the geometric cycle consistency loss.
\begin{equation}
    \mathcal{L}_{\text{rec}}^{\text{total}} =  \alpha (\mathcal{L}_{\text{I}} + \mathcal{L}_{\text{M}}) + \beta \mathcal{L}_{\text{G}}
\end{equation}
\subsection{Pose Cycle Consistency Loss}
The projection based losses form weak supervisory signals to train the pose prediction network. 
While there is no direct pose information available for the input images, 
the projected images and corresponding pose pairs $\{\hat{I}^{i}, \mathbf{v}^{i}\}_{1}^{k}$ can be considered as pseudo ground-truth pairs for training the pose network.
We input the image projections $\{\hat{I}^{i}\}_{1}^{k}$ to the pose prediction network $\mathcal{N}_{\text{pose}}$ to obtain the corresponding pose predictions $\{\mathbf{\hat{v}}^{i}\}_{1}^{k}$ (Fig~\ref{fig:net_arch}). The corresponding viewpoints $\{\mathbf{v}^{i}\}_{1}^{k}$ are then used as ground-truth to train $\mathcal{N}_{\text{pose}}$. The pose loss is obtained as follows:
\begin{equation}
    \mathcal{L}_{\text{pose}} = \frac{1}{k}\sum\limits_{i=1}^{k}|\mathbf{v}^i-\hat{\mathbf{v}}^i|
\end{equation}

The final training objective for the pose network is a combination of pose cycle consistency loss and image and mask losses (Eq.~\ref{eq:rgb_loss} and ~\ref{eq:bce_loss}). This ensures that the pose loss is dependent on the pose predictions of the input image, while simultaneously being optimized with a stronger supervision using the projected images.
\begin{equation}
    \mathcal{L}_{\text{pose}}^{\text{total}} =  \gamma (\mathcal{L}_{\text{I}} + \mathcal{L}_{\text{M}}) + \rho \mathcal{L}_{\text{pose}}
\end{equation}

\begin{figure}
    \centering
    \includegraphics[width=\linewidth]{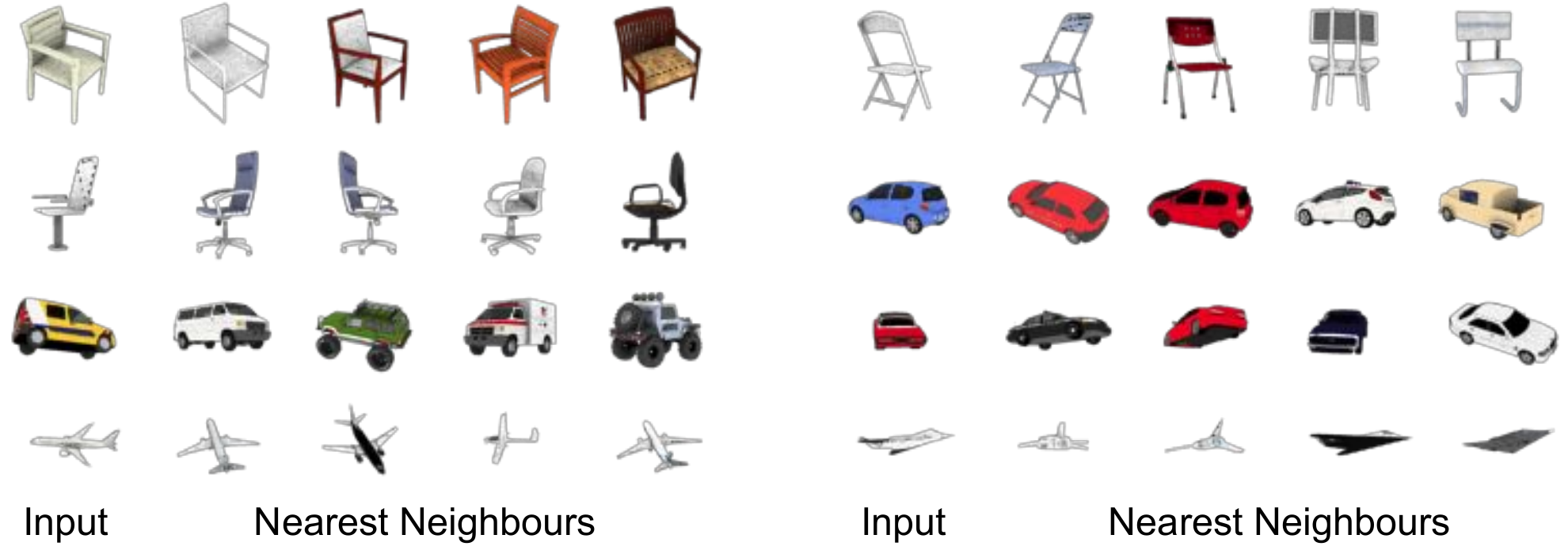}
    \caption{\textbf{Sample k-nearest neighbours.} We utilize our single-view trained reconstruction network to obtain k-nearest neighbour samples from the train set. Note that the neighbours have different poses and have different color distribution, but have similar 3D shape which provides us with additional information on the geometry of the object.} 
    \label{fig:knn}
\end{figure}

\subsection{Nearest Neighbours Consistency Loss}
Earlier works~\cite{navaneet2019capnet} demonstrate that even just a single additional view as supervision during training significantly improves the reconstruction quality.
However, as mentioned previously, assuming the presence of such multi-view images during training curtails the practical utility and prevents the applicability on real-world single image datasets.
In order to remain in the constrained setting, but improve reconstructions with the use of multiple image supervision, we propose mining images from the training set which belong to similar 3D models. For every input image, we find the closest neighbours such that they have similar underlying 3D shapes, and use projection consistency based loss, termed `nearest neighbours consistency loss', to assist the training of the network. To find the nearest neighbours in the 3D domain in a self-supervised fashion, we need features which embed the 3D shape information. Utilizing features from networks trained on 2D tasks (for \eg{classification on ImageNet dataset}), would provide neighbours which are similar in color and viewpoint, but not necessarily in 3D shape. 
Alternatively, to quantify the 3D similarity, we consider the encoded features of our proposed reconstruction network. 
Nearest neighbours from training set are obtained by comparing the Euclidean distances in the encoded feature space. Sample nearest neighbour images are shown in Fig.~\ref{fig:knn}. We observe that the retrievals are similar in shape and have diversity in terms of pose and color. 
During training, nearest neighbours of the input image are utilized as additional supervision. The neighbour images are passed through $\mathcal{N}_\text{pose}$ to obtain the corresponding poses. The reconstructed point cloud obtained from the input image is projected from these viewpoints. We then enforce silhouette loss in Eq.~\ref{eq:bce_loss} on these projections using the ground-truth silhouettes of the neighbour images. This is possible since the geometry of the input and the neighbours are similar and thus the projections from the input model closely match those from the neighbours. Note that the loss is enforced using only masks and not the color images since the neighbours might have different color distribution. The mask losses are summed over $n$ neighbours to get the total nearest neighbours loss. This is used in addition to the losses mentioned in Eq.~\ref{eq:rgb_loss} and ~\ref{eq:bce_loss} to train the reconstruction network. 
\begin{equation}
    \mathcal{L}_{\text{NN}} = \sum \limits_{i=1}^{n} \mathcal{L}^{i}_{\text{M}}
    \label{eq:knn}
\end{equation}

\subsection{Symmetry Loss}
Since all the object categories we consider in our experiments have a minimum of one plane of symmetry, we further regularize the network to obtain symmetric 
reconstructions with respect to a pre-defined plane. Without loss of generality, let us assume that the point clouds are symmetric with respect to the 
xz-plane. Then, the symmetry loss is given by:
\begin{equation}
    \mathcal{L}_{\text{sym}} = \text{d}_{\text{Ch}}(\hat{P}^{+}, \hat{P}^{-})
\end{equation}
where $\hat{P}^{+}$ is the set of points in $\hat{P}$ with positive $y$ values and $\hat{P}^{-}$ is the reflection about the xz-plane of the points in 
$\hat{P}$ with negative $y$ values. The symmetry loss helps in obtaining improved geometry of reconstructions consistent with the ground truth and avoids
overfitting to the input image. Due to the absence of ground truth pose values, the co-ordinate system for the predicted camera poses is not pre-determined. The choice of plane of symmetry in enforcing symmetry loss can also help align the reconstructions to a predefined canonical pose. 
The total reconstruction loss with nearest neighbours and symmetry losses is as follows:
\begin{equation}
    \mathcal{L}_{\text{rec}}^{\text{total}} = \alpha (\mathcal{L}_{\text{I}} + \mathcal{L}_{\text{M}}) + \beta \mathcal{L}_{\text{G}} + \eta \mathcal{L}_{\text{NN}}
                                + \kappa \mathcal{L}_{\text{sym}}
    \label{eq:loss_total}
\end{equation}

\subsection{Inference Stage Optimization (ISO)}
Our self-supervised approach, which relies only on the input images and corresponding object silhouettes for training, is ideally poised for instance specific optimization during inference. At inference, we predict both the 3D point locations and the input image viewpoint. To obtain highly corresponding reconstructions, we aim to minimize the difference between the input and the projected image (from predicted viewpoint) during inference. To ensure that the reconstructions are not degraded in the regions occluded in the input image, we employ additional regularization.
Note that while CAPNet~\cite{navaneet2019capnet} too performs inference stage optimization, unlike our work, the authors assume known viewpoint.
The regularization loss formulation is as follows: 
\begin{equation}
    \mathcal{L}_{\text{reg}} = \text{d}_{\text{ch}}(\hat{P}, \hat{P}_{\text{O}})
\end{equation}
where $\hat{P}$ and $\hat{P}_{\text{O}}$ are the initial and optimized point clouds.
We also use the symmetry loss as an additional form of regularization to enable the network to optimize for the regions in the point cloud visible in 
the input image while suitably modifying the points corresponding to the occluded regions. The objective function during ISO is given by:
\begin{equation}
    \mathcal{L}_{\text{ISO}} = \alpha (\mathcal{L}_{\text{I}} + \mathcal{L}_{\text{M}}) + \lambda(\mathcal{L}_{\text{reg}}) + \kappa (\mathcal{L}_{\text{sym}})
\end{equation}
\section{Experiments}

\subsection{Implementation Details}
We use a two-branch network to simultaneously obtain shape and color reconstructions. Separate models are used for training on each object category. The number of projections, $k$ is set to four and the number of points in reconstructed point cloud to $1024$. 
Adam optimizer with a learning rate of $0.00005$ is used for training the network. The hyperparameters $\alpha$, $\beta$, $\gamma$ and $\rho$ are set to $100$, $10^4$, $1$ and $1$ respectively. Architecture details, additional details on hyperparameter settings and training schedules are provided in the supplementary material. We publicly release the code.\footnote{Code is available at \url{https://github.com/val-iisc/ssl\_3d\_recon}}

\subsection{Datasets}
\noindent \textbf{ShapeNet}~\cite{chang2015shapenet}: ShapeNet is a curated set of synthetic 3D mesh models. We sample points on the surface of the meshes to obtain the corresponding point clouds for evaluation. To create the set of input images, we render the mesh models from a single random view per object instance. All the experiments are performed on the representative car, chair and airplane (denoted as aero) categories. \\
\noindent \textbf{Pix3D}~\cite{pix3d}: Pix3D is a repository of aligned real-world image and 3D model pairs. The dataset exhibits great diversity in terms of object shapes and backgrounds and is highly challenging. We consider the chair category of Pix3D in our experiments. Since the dataset is small, we only perform evaluation on the Pix3D dataset. \\
We use the train/val/test splits provided by DIFFER~\cite{navaneet2019differ} in all our experiments. For ease of comparison, all the Chamfer and EMD metrics are scaled by $100$.

\subsection{Evaluation Methodology}
Since point clouds are unordered representations, we use Chamfer distance and earth mover's distance (EMD) to evaluate the reconstructions. For evaluation, we randomly sample $1024$ points from the reconstructions if they contain higher number of points. The Chamfer distance between two point clouds $P$ and $\hat{P}$ is defined as $d_{\text{Chamfer}}(P,\hat{P}) = \sum_{x\in P}\min_{y\in \hat{P}}{||x-y||}^2_2 + \sum_{x\in \hat{P}}\min_{y\in P}{||x-y||}^2_2$. EMD between two point clouds is defined as $d_{\text{EMD}}(P,\hat{P})=\min_{\phi:P\rightarrow \hat{P}}\sum_{\alpha\in P}||\alpha-\phi(\alpha)||_2$ where $\phi(\cdot)$ is a bijection from $P$ to $\hat{P}$. For the pose unsupervised approaches, the models are aligned using a global rotation matrix obtained by minimizing the Chamfer error on the validation set. To evaluate color metrics, we project each reconstruction from $10$ randomly sampled viewpoints and compute the $\mathcal{L}_2$ distance using the ground-truth images. We report the median angular error and accuracy in the pose prediction evaluation. 
In addition, the pose metrics are also calculated by utilizing the ground truth orientation. The predicted point cloud is `flipped' (rotated by 180\degree) if the error is more than 90\degree.

\subsection{Baseline Approaches}
We compare the proposed approach with two state-of-the-art approaches on 2D supervised single image based 3D point cloud reconstruction. Specifically, we use the following variants of the works:\\
\noindent \textbf{DIFFER:} DIFFER~\cite{navaneet2019differ} proposed a differentiable module to project point cloud features on to the 2D plane, which enables it to utilize input images for training. Note that DIFFER utilizes ground truth pose values for the input image and hence has a higher degree of supervision compared to our approach. Codes and settings provided by the authors are used to train the network. \\
\noindent \textbf{ULSP:} Insafutdinov \etal~\cite{insafutdinov2018pointclouds} proposed a multi-view consistency based unsupervised approach for point cloud reconstruction. While the approach does not make use of ground truth pose values, it requires multiple images and their corresponding foreground masks from different viewpoints per 3D object instance. Hence, the work is not directly comparable to our approach which uses just a single image per model. To remain as close as possible to this setting, we train ULSP with supervision from two views per model using the code provided by the authors. \\
\noindent \textbf{ULSP\_Sup:} We consider a variant of ULSP~\cite{insafutdinov2018pointclouds} with ground truth camera pose supervision. Similar to DIFFER, this is trained with one input viewpoint per 3D model.\\

We also provide comparison with two variants of the proposed approach - `Ours-CC' and `Ours-NN'. Ours-CC is trained only with the cycle consistency losses while NN consistency loss is used in addition in Ours-NN.  

\subsection{Effect of Cycle Consistency Losses}

We first analyze the role of the proposed consistency losses in improving the reconstructions in a self-supervised setting  (Table~\ref{tab:const_ablation}). In the absence of both $\mathcal{L}_{\text{G}}$ and $\mathcal{L}_{\text{pose}}$ (Ours-No-CC), the network fails to learn meaningful 3D reconstructions. When both the cyclic losses are employed (Ours-CC), we observe that the network learns the underlying 3D shapes of the objects and thus results in effective reconstructions. We present detailed ablations for individual loss components in the supplementary material. 

\begin{table}
    \centering
    \scalebox{0.9}{
    \begin{tabular}{llllllll}
    \toprule
    \multirow{2}{*}{Method}    & \multicolumn{3}{c}{Chamfer}    & \multicolumn{3}{c}{EMD} \\
     & Car & Chair & Aero & Car & Chair & Aero \\
    \midrule
                             Ours-No-CC   & 10.33	& 21.84   & 15.06  & 18.32	& 23.40  & 16.12 \\
                             Ours-CC   & \textbf{6.39}	& \textbf{13.58}   &  \textbf{8.66}  & \textbf{6.42}   & \textbf{16.46} & \textbf{12.53} \\
    \bottomrule
    \end{tabular}}\\
    \vspace{5pt}\caption{\textbf{Effect of Consistency Loss.} We evaluate the effect of the proposed consistency losses on reconstruction metrics. The network fails to train in the absence of the consistency losses in the self-supervised setting.} 
    \label{tab:const_ablation}
\end{table}

We also demonstrate the utility of the proposed geometric losses in the pose supervised setting for single image based 3D reconstructions. Specifically, we use the proposed loss $\mathcal{L}_\text{G}$ atop pose supervised DIFFER and ULSP\_Sup to optimize the corresponding reconstruction networks. 
Table~\ref{tab:geometric_loss_supervised} suggests that geometric loss can significantly improve the performance of existing supervised approaches as well. 

\begin{table}
    \centering
    \scalebox{0.9}{
    \begin{tabular}{lllllll}
        \toprule
        \multirow{2}{*}{Method} & \multicolumn{3}{c}{Chamfer}    & \multicolumn{3}{c}{EMD} \\
                 & Car & Chair  & Aero & Car & Chair & Aero \\
        \midrule
        DIFFER                          & 6.35	& 9.78  & 5.67 & 6.03	& 16.21  & 9.9 \\
        DIFFER + $\mathcal{L}_G$        & \textbf{5.63}	& \textbf{9.23}  &\textbf{5.58} & \textbf{5.35}	& \textbf{13.07} &\textbf{9.44} \\ \midrule
        ULSP\_Sup                       & 6.64  & 10.49	& \textbf{5.70} & 6.89 & 10.93 & \textbf{7.43}\\
        ULSP\_Sup + $\mathcal{L}_G$     & \textbf{6.13}  &  \textbf{10.0}  & 7.37 & \textbf{5.83}  & \textbf{10.24} & 9.99\\
        \bottomrule
    \end{tabular}}
    \vspace{5pt}\caption{\textbf{Portability of Geometric Consistency.} 
    Using our geometric consistency loss 
    atop supervised approaches results in significant gains in reconstruction performance.}
    \label{tab:geometric_loss_supervised}
\end{table}
\begin{figure*}
    \centering
    \includegraphics[width=\linewidth]{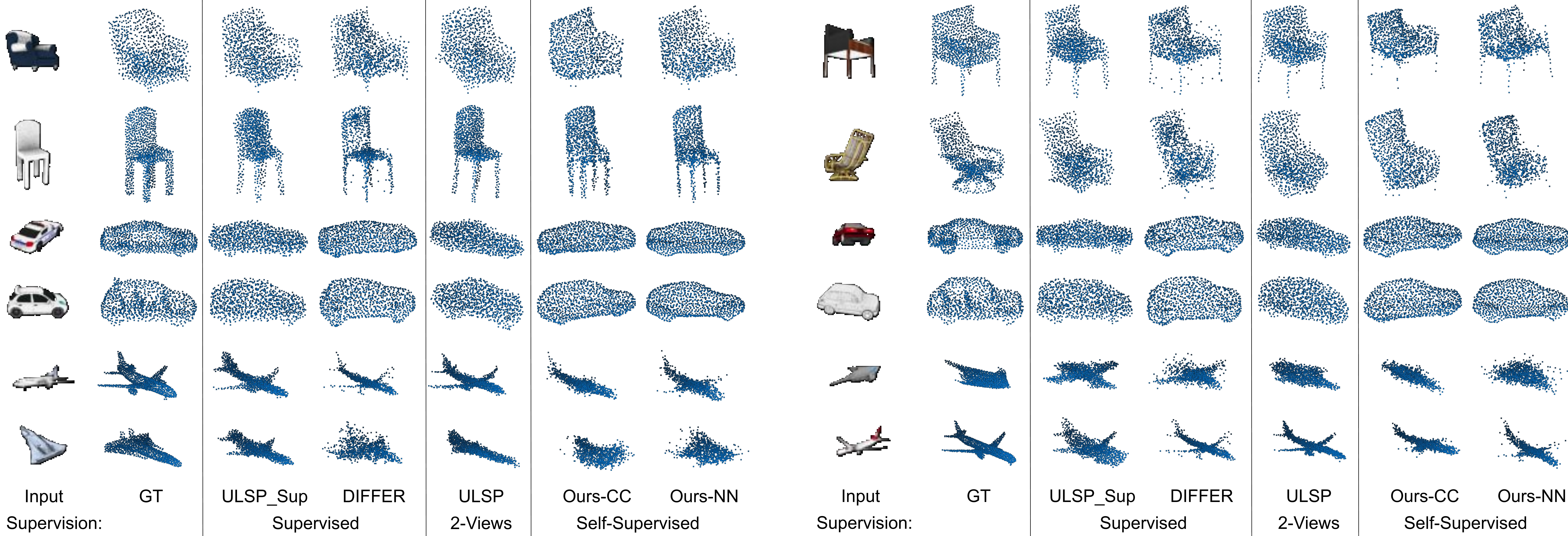}
    \caption{\textbf{Comparisons on ShapeNet.} We provide comparison with both pose and multi-view supervised approaches on ShapeNet. Our approach is on par with the supervised approaches in terms of correspondence of the reconstruction to the input image. Our car reconstructions have significantly better shape and uniformity in points compared to the supervised approaches.}
    \label{fig:shapenet_comparison}
\end{figure*}

\subsection{Reconstruction Results}
Quantitative and qualitative comparisons of the proposed self-supervised approach with other multi-view and pose supervised approaches on the ShapeNet dataset are provided in Table~\ref{tab:recon_main} and Fig.~\ref{fig:shapenet_comparison} respectively. The performance of our approach is comparable to those utilizing higher levels of supervision. 
For the baseline approaches, we observe that pose supervised ULSP\_Sup is marginally better than the two-views supervised ULSP in the case of chairs and 
airplanes and significantly better in the case of cars. Our car reconstruction metric is close to the supervised ULSP network and is better than other approaches. Notably, while we use the same projection module and projection consistency losses as in DIFFER, we outperform the pose supervised DIFFER in most of the quantitative metrics. This demonstrates the utility of the additional cycle and nearest neighbour consistency loss for reconstruction and pose prediction. The addition of nearest neighbour significantly boosts the reconstruction performance, particularly in the case of the more challenging chair category. In the car and airplane categories, there is apparent visual improvement in the shape and spread of points with the use of nearest neighbours. While we are able to effectively capture the geometry of the object, points are sparsely distributed in the thin regions such as legs in the case of chairs. However, we can observe similar sparse point distributions in the case of DIFFER~\cite{navaneet2019differ}. We also present qualitative (Fig.~\ref{fig:tso}) and quantitative (in supplementary) results on inference stage optimization. The reconstructions have greater correspondence with the input image as observed in the silhouettes before and after optimization in Fig.~\ref{fig:tso}. Reconstruction metrics indicate that the point clouds are preserved in regions not observed in the test input. 
Additional qualitative results, ablations on symmetry and nearest neighbours consistency loss and failure cases are provided in the supplementary.

To show the adaptability of our approach to real-world datasets, we evaluate it on the Pix3D dataset. Note that since the dataset consists of very few models, we perform evaluation of the networks trained on ShapeNet dataset. For synthetic to real domain adaptation, we train on ShapeNet dataset with the input images overlaid with random natural scene backgrounds. Our approach performs comparably to the pose supervised DIFFER approach both quantitatively (Table~\ref{tab:pix3d}) and qualitatively (Fig.~\ref{fig:pix3d}).

Fig.~\ref{fig:color_metrics} presents qualitative results on color prediction on ShapeNet dataset. For effective evaluation, we project each ground truth and predicted model from $10$ randomly sampled viewpoints and calculate the channel-wise $\mathcal{L}_2$ loss between them. Our reconstructions result in greater visual correspondence with the input image, particularly in the case of cars. Quantitative results are provided in 
the supplementary.   
\begin{table*}
    \centering
    \scalebox{0.9}{
    \begin{tabular}{lllllllllll}
    \toprule
    \multirow{2}{*}{Method}   & \multirow{2}{*}{Pose}  & \multirow{2}{*}{Views}   & \multicolumn{4}{c}{Chamfer}    & \multicolumn{4}{c}{EMD} \\
    & & & Car & Chair & Aero & Mean & Car & Chair & Aero & Mean\\
    \midrule
    ULSP\_Sup                     & Yes  & 1 view    & 5.4   & 9.72  & 5.91 & 7.01  & 4.78  & 10.18  & 7.66  & 7.54  \\
    DIFFER                       & Yes  & 1 view     & 6.35  & 9.78  & 5.67 & 7.27  & 6.03  & 16.21  & 9.90  & 10.71  \\ \midrule
    ULSP                         & No   & 2 views    & 7.02  & 9.87  & 5.96 & 7.62  & 7.99  & 10.56  & 8.06  & 8.87  \\ \midrule
    Ours-CC                         & No   & 1 view  & 6.39  & 13.58 & 8.66 & 9.54  & 6.42  & 16.46  &12.53  & 11.8  \\
    Ours-NN                         & No   & 1 view  & 5.48  & 10.91 & 7.11 & 7.83  & 4.95  & 14.93  &11.07  & 10.31  \\
    \bottomrule
    \end{tabular}}
    \vspace{5pt}\caption{\textbf{Reconstruction Metrics on ShapeNet.} Despite being self-supervised, lacking the input pose values and with just the input image as supervision, we perform comparably to or even outperform other state-of-the-art approaches requiring higher degree of supervision.} 
    \label{tab:recon_main}
\end{table*}
\begin{figure}
    \centering
    \includegraphics[width=\linewidth]{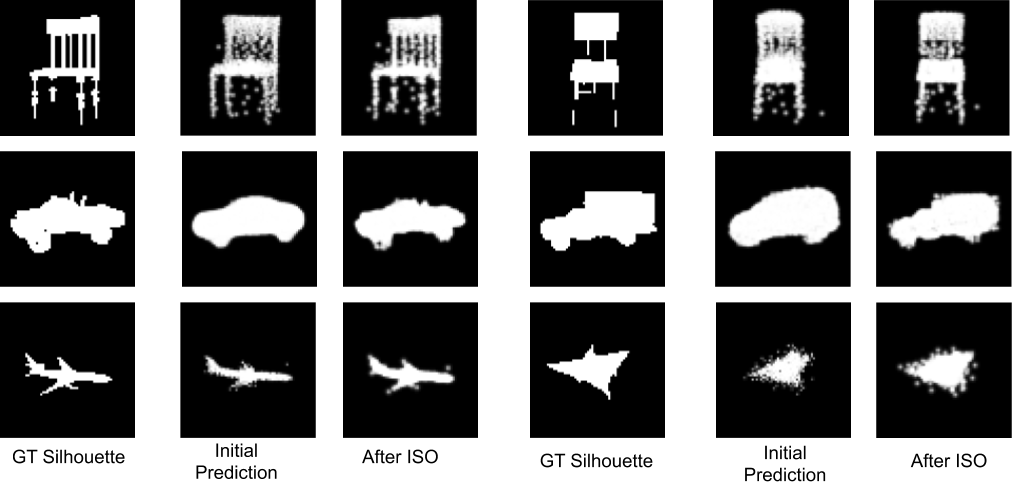}
    \caption{\textbf{Inference stage optimization (ISO).} Optimization during inference results in greater correspondence to the input image. Regularization is employed to maintain the shape in regions occluded in the input image.}
    \label{fig:tso}
\end{figure}
\begin{figure}
    \centering
    \includegraphics[width=\linewidth]{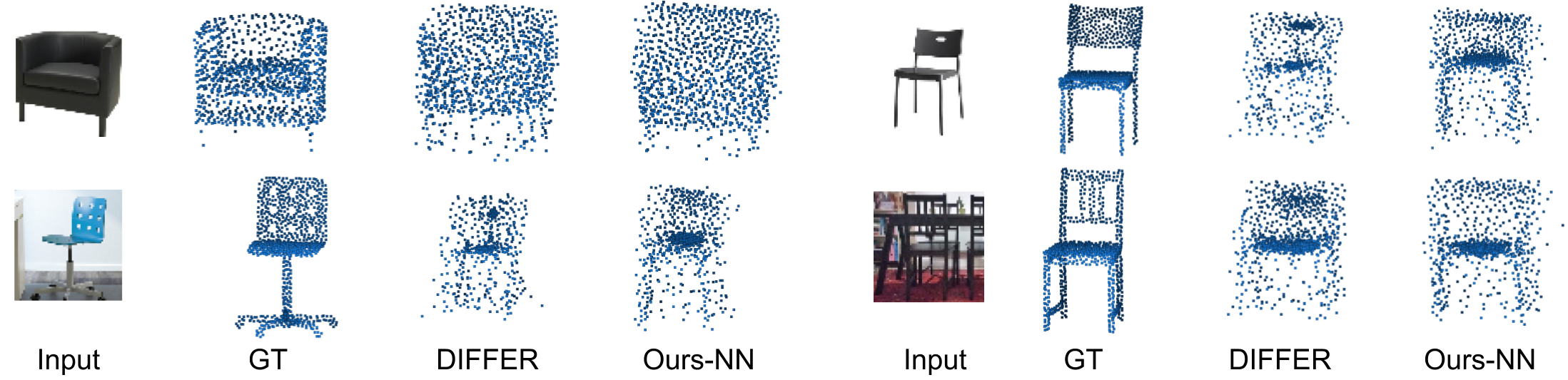}
    \caption{\textbf{Comparisons on Pix3D.} Since both DIFFER and the proposed approach are trained on ShapeNet and evaluated on Pix3D, the correspondence to input in reconstructions is lower compared to that on ShapeNet. However, our reconstructions have marginally better shape and point spread compared to the supervised DIFFER approach.}
    \label{fig:pix3d}
\end{figure}

\begin{table}
    \centering
    \scalebox{0.9}{
    \begin{tabular}{lll}
        \toprule
        Method  & Chamfer & EMD \\
        \midrule
        DIFFER      & 14.33	    & 16.09  \\
        Ours-NN     & 14.52	    & 15.82  \\
        \bottomrule
    \end{tabular}}
    \vspace{5pt}\caption{\textbf{Reconstruction Results on Pix3D.} We evaluate both the pose supervised DIFFER and our approach on the real-world Pix3D~\cite{pix3d} dataset. Our self-supervised approach performs comparably to the pose supervised one and adapts well to the real-world dataset.} 
    \label{tab:pix3d}
\end{table}
\begin{figure}
    \centering
    \includegraphics[width=1.\linewidth]{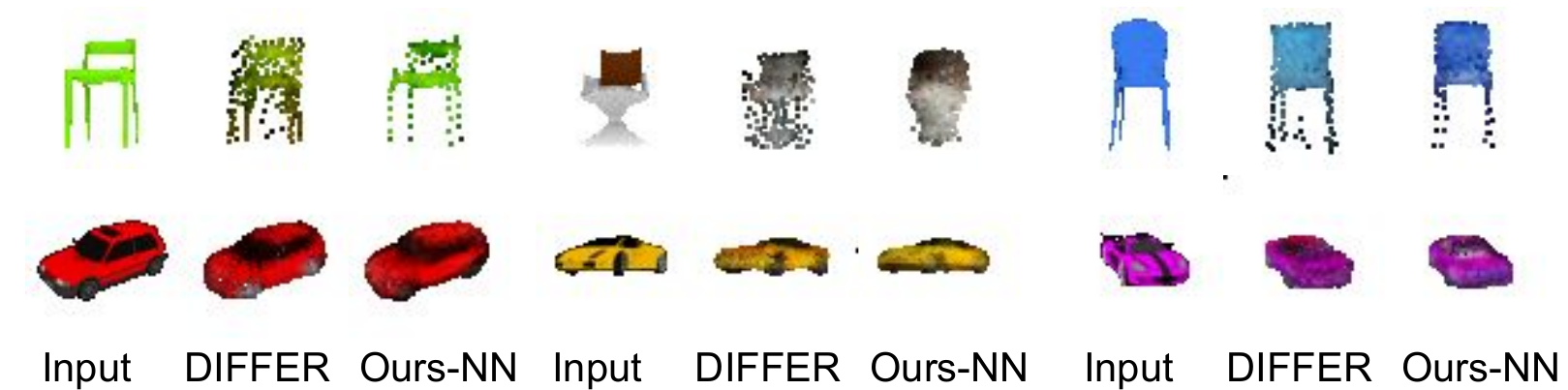}
    \caption{\textbf{2D Color Projections.} Our colored projections have greater visual correspondence with the input images compared to the supervised DIFFER approach.}
    \label{fig:color_metrics}
\end{figure}
\subsection{Pose Prediction Results}
Table~\ref{tab:pose_metrics} presents median error and accuracy of our pose prediction network. We report results both with (`flip') and without (`No-flip') the use of ground-truth orientation. Ours-CC achieves high accuracy on the car category. However, in the chair category, where there exists higher ambiguity, Ours-CC performs significantly worse. 
Due to the existence of multiple planes of symmetry in certain airplane models, the network often 
predicts the incorrect orientation, as observed in the high median error. But when the ground truth orientation is used to calculate the metrics, such conflicts are resolved leading to significantly better metrics. In all the categories, we observe that the pose metrics reliably improve upon the introduction of nearest neighbor consistency ($\mathcal{L}_\text{NN}$), further highlighting the need for such a loss. We also observe that the pose and reconstruction metrics are correlated and thus incorrect prediction in either of them significantly affects the other.  
\begin{table}
    \centering
    \scalebox{0.9}{
    \begin{tabular}{llllll}
    \toprule
    \multirow{2}{*}{Categ.}   &\multirow{2}{*}{Method}   & \multicolumn{2}{c}{Median Error}  & \multicolumn{2}{c}{Accuracy}\\ 
        &   & No-flip     & Flip   & No-flip     & Flip \\
    \midrule
    \multirow{2}{*}{Car}     & Ours-CC    & 7.58	    & \textbf{5.54}     & 74.07	    & 94.4  \\
                             & Ours-NN       & \textbf{6.85}	    & 5.55	    & \textbf{75.87}	    & \textbf{93.4}  \\ \midrule
    \multirow{2}{*}{Chair}     & Ours-CC  & 41.86	& 33.78	    & 41.45	    & 45.72  \\
                             & Ours-NN       & \textbf{19.69}	& \textbf{17.79}	    & \textbf{59.14}	    & \textbf{64.16}  \\ \midrule
    \multirow{2}{*}{Aero}     & Ours-CC    & 88.29	    & 38.53	    & 	20.99    & 40.74  \\
                             & Ours-NN         & \textbf{43.36}	    & \textbf{19.52}	    & \textbf{42.34}	    & \textbf{60.74}  \\ 
    \bottomrule
    \end{tabular}}
    \vspace{5pt}\caption{\textbf{Pose Metrics on ShapeNet.}  
    Pose metrics are remarkably good for the car category for both our approaches. In the challenging chair and airplane categories, use of nearest neighbours (Ours-NN) significantly boosts the predictions.} 
    \label{tab:pose_metrics}
\end{table}
\subsection{Point Correspondences and Part Transfer}

\begin{figure}
    \centering
    \includegraphics[width=1.\linewidth]{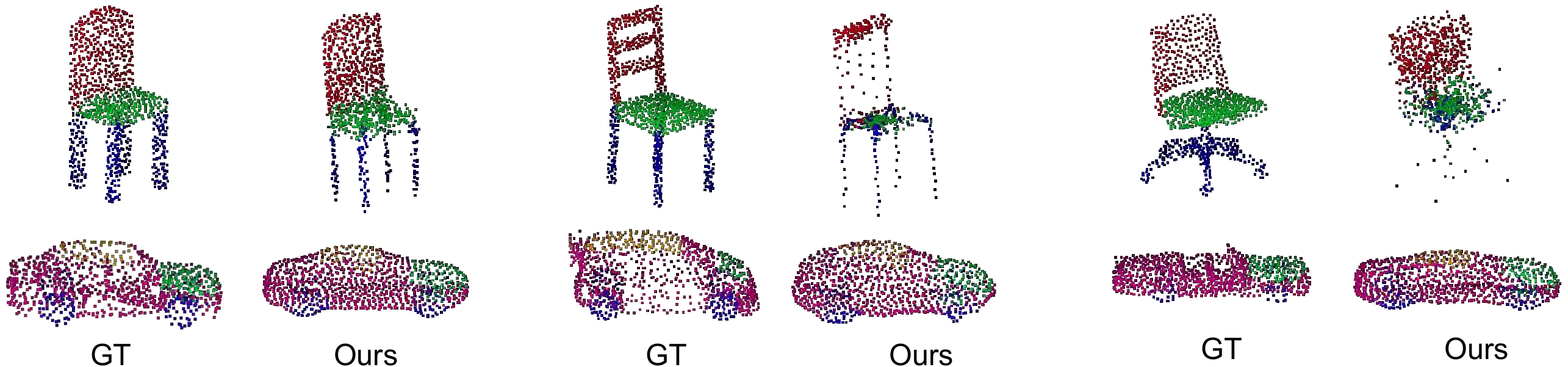}
    \caption{\textbf{Part Transfer.} Semantic part segmented ground truth and reconstructed point clouds. Correspondences between the reconstructed point clouds are used for consistent part segmentation transfer across models.}
    \label{fig:part_seg}
\end{figure}

In our reconstructions, we observe that points with similar indices in the regressed points have spatial correspondence even though we do not explicitly enforce it. We use a colored UV map to visualize the point correspondences (see Supplementary for more details). We utilize this correspondence for the task of single-shot semantic part segmentation. We use a single ground-truth part-segmented model to transfer part labels across all models based on point indices. Results (Fig.~\ref{fig:part_seg}) indicate that our network is effective in obtaining 3D part segmentations using just a single ground-truth model.
\vspace{-0mm}
\section{Conclusion}
We propose a self-supervised approach for single image based 3D point cloud reconstruction. We develop novel geometric and pose cycle consistency losses to effectively train our reconstruction and pose networks in a self-supervised manner. Through the use of training images with similar 3D shape, we mimic the effect of training with multi-view supervision using a single-view dataset. We benchmark our reconstruction, color and pose prediction networks on the ShapeNet dataset, achieving comparable performance to pose and multi-view supervised approaches. The role of all the proposed losses is thoroughly analyzed. We further demonstrate the utility of our approach through reconstruction results on the real-world Pix3D dataset and qualitative results on possible applications like dense point correspondence and 3D part segmentation. In future, we like to address the issue of sparse point predictions in thin structures and further improve the reconstruction quality.    
\paragraph{Acknowledgement}
This work was supported by Sudha Murthy Chair Project, Pratiksha Trust, IISc.

{\small
\bibliographystyle{ieee_fullname}
\bibliography{bibliography}
}

\newpage

\makeatletter
\renewcommand\AB@affilsepx{, \protect\Affilfont}
\makeatother

\title{From Image Collections to Point Clouds with Self-supervised \\Shape and Pose Networks: Supplementary Material}

\author{\vspace{-1em}%
K L Navaneet$^1$\quad 
Ansu Mathew$^1$\quad 
Shashank Kashyap$^1$\quad 
Wei-Chih Hung$^2$\\ \vspace{0.0em}%
Varun Jampani$^3$\quad 
R. Venkatesh Babu$^1$\\ \vspace{1em}%
$^1$Indian Institute of Science \qquad
$^2$University of California, Merced \qquad 
$^3$Google Research\\
}
\date{}
\maketitle
\thispagestyle{empty}
\setcounter{section}{0}
\setcounter{table}{0}
\setcounter{figure}{0}
\renewcommand{\thetable}{\arabic{table}}
\renewcommand\thefigure{\arabic{figure}}
\renewcommand{\theHtable}{Supplement.\thetable}
\renewcommand{\theHfigure}{Supplement.\thefigure}
The supplementary document is arranged as follows: We provide training schedule and additional implementation details in the initial sections. We add more detailed ablations on the role of individual components of the proposed cycle consistency based losses. Subsequently we present experimental results on the effect of number of nearest neighbours, consistency and symmetry losses, dense point correspondence, inference stage optimization and colored point cloud reconstruction. We provide qualitative 
results on failure modes of our approach. Lastly, we provide the architectural details of our reconstruction and pose prediction networks.\footnote{Code is available at \url{https://github.com/val-iisc/ssl_3d_recon}} 

\section{Training Schedule}
We train our networks for $400000$ iterations using Adam optimizer with a learning rate of 0.0005. For training our approach, we observe that the pose prediction network converges at a much earlier stage compared to the reconstruction network. At the half-way stage ($200000$ iterations), we freeze the pose network and train the reconstruction network with just image and mask losses, similar to the DIFFER baseline. We observe that this helps in obtaining better 3D shape reconstructions and eliminate outlier points in predictions. 

\section{Additional Implementation Details}
We choose the optimal hyperparameter values based on the reconstruction performance on the validation set. The weight for geometric consistency loss, $\beta$ is set to 10000 and pose consistency loss, $\rho$ is set to $1$. The weight for nearest neighbours consistency loss $\kappa$ is set to be same as that for mask loss $\alpha$. During the second half of the training schedule, the weights for consistency losses $\beta$ and $\rho$ are set to $0$ and that of image and mask losses $\alpha$ is reduced to $10$.
In the experiments on nearest neighbours, we consider five nearest neighbours for every input among which $n$ images are sampled randomly. The effect of the number of neighbours chosen, $n$, is presented in Fig.~\ref{fig:Ablation study on Nearest Neighbors} and Table~\ref{tab:nearest_neighbours_comparison}. In inference stage optimization experiments, the weights for regularization and symmetry loss, $\lambda$ and $\kappa$ are both set to $500$.

\section{Role of Cycle Consistency Losses}
We present quantitative ablation on the role of individual components of our proposed cycle consistency loss in Table~\ref{tab:consistency_ablation_suppl}. We present qualitative comparison of reconstruction with and without these losses in a self-supervised setting in Fig.~\ref{fig:consistency_loss}. The network fails to learn meaningful 3D shapes in the absence of the proposed losses, while the reconstructions closely match the input when the losses are utilized. We also observe that each of the individual losses help improve the reconstructions and the best performance is obtained when all the losses are combined. 
Fig.~\ref{fig:ULSP_geometric}, displays the qualitative results on the effect of geometric consistency loss on the pose supervised ULSP\_Sup approach. We observe a significant improvement in the reconstruction quality, suggesting the portable nature of the proposed loss. 
\begin{table*}
    \centering
    \scalebox{0.9}{
    \begin{tabular}{ccclllllll}
    \toprule
    \multirow{2}{*}{Geometric CC}   & \multirow{2}{*}{Pose CC} & Nearest  &\multicolumn{3}{c}{Chamfer}    & \multicolumn{3}{c}{EMD} \\
      &  & Neighbor CC & Car & Chair & Aero & Car & Chair & Aero \\
    \midrule
                             \xmark & \xmark  & \xmark      & 10.33	    & 21.84   & 15.06  & 18.32	& 23.40  & 16.12 \\
                             \cmark & \xmark & \xmark   &5.78 &27.89&10.77 & 7.07 &26.9 &15.76   \\
                             \xmark & \cmark & \xmark    &11.31 &11.46& 12.47 &11.59 &14.97&15.26  \\
                             \cmark & \cmark & \xmark     & 6.39  & 13.58 & 8.66 & 6.42  & 16.46  &12.53 \\
                             \cmark & \cmark & \cmark     & \textbf{5.48}  & \textbf{10.91} & \textbf{7.11} & \textbf{4.95}  & \textbf{14.93}  & \textbf{11.07} \\
    \bottomrule
    \end{tabular}}
    \vspace{3mm}
    \caption{\textbf{Effect of Consistency Loss.} We evaluate the effect of the proposed consistency losses on reconstruction metrics. The network fails to train in the absence of the consistency losses in the self-supervised setting. Each of the proposed losses is necessary to obtain the optimal performance.} 
    \label{tab:consistency_ablation_suppl}
\end{table*}

\section{Effect of Nearest Neighbours}

\begin{figure*}[h]
    \centering
        \includegraphics[width=.9\linewidth]{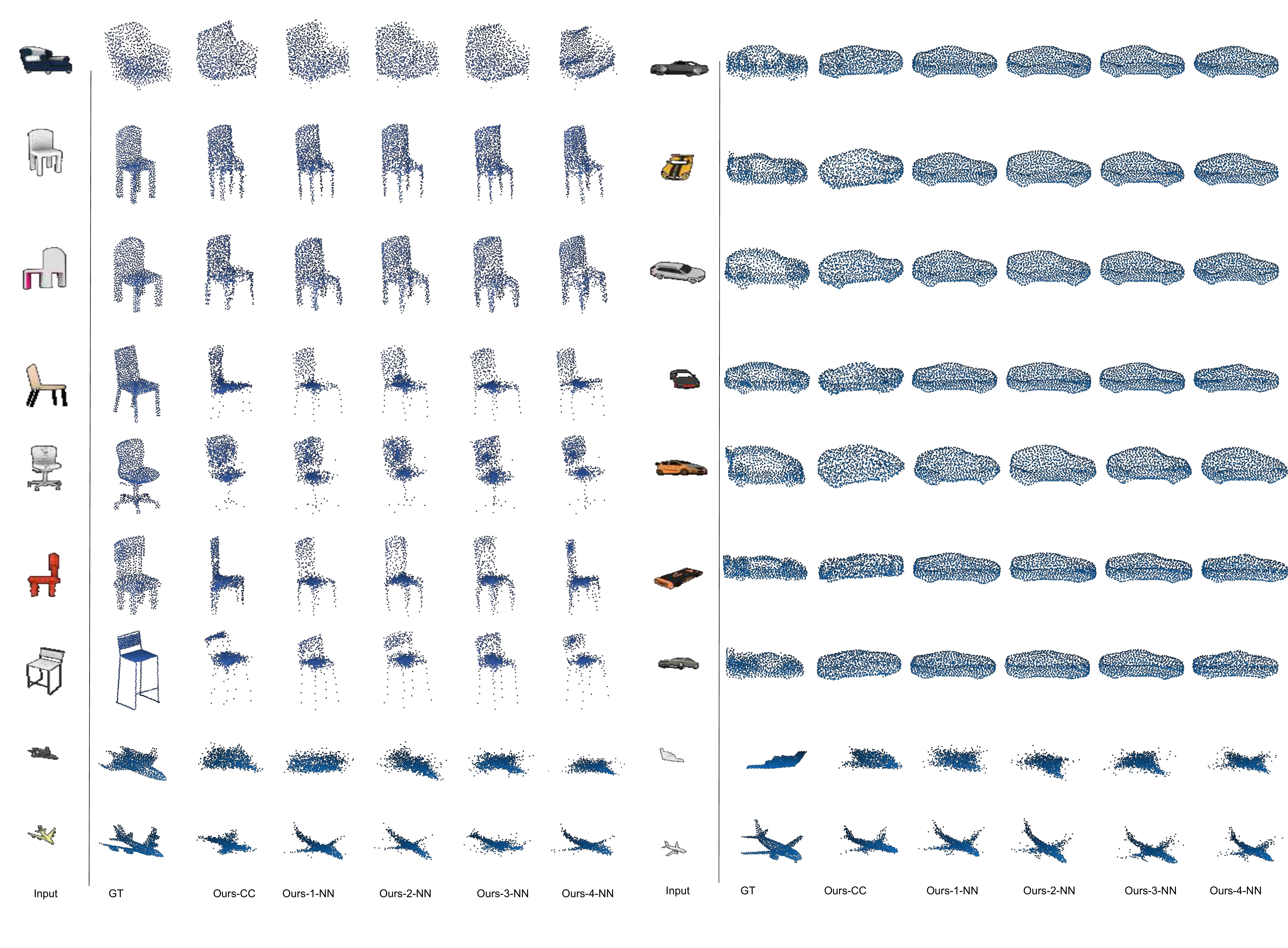}
    \caption{\textbf{Effect of Nearest Neighbours.} We examine the effect of number of images of nearest neighbours on the reconstruction metrics. The performance improves or remains nearly same as more number of images are considered. The best performance is achieved when one or two images are used while reconstructions suffer when more than 3 images are utilized.}
    \label{fig:Ablation study on Nearest Neighbors}
\end{figure*}

\begin{figure*}
    \centering
        \includegraphics[width=0.9\linewidth]{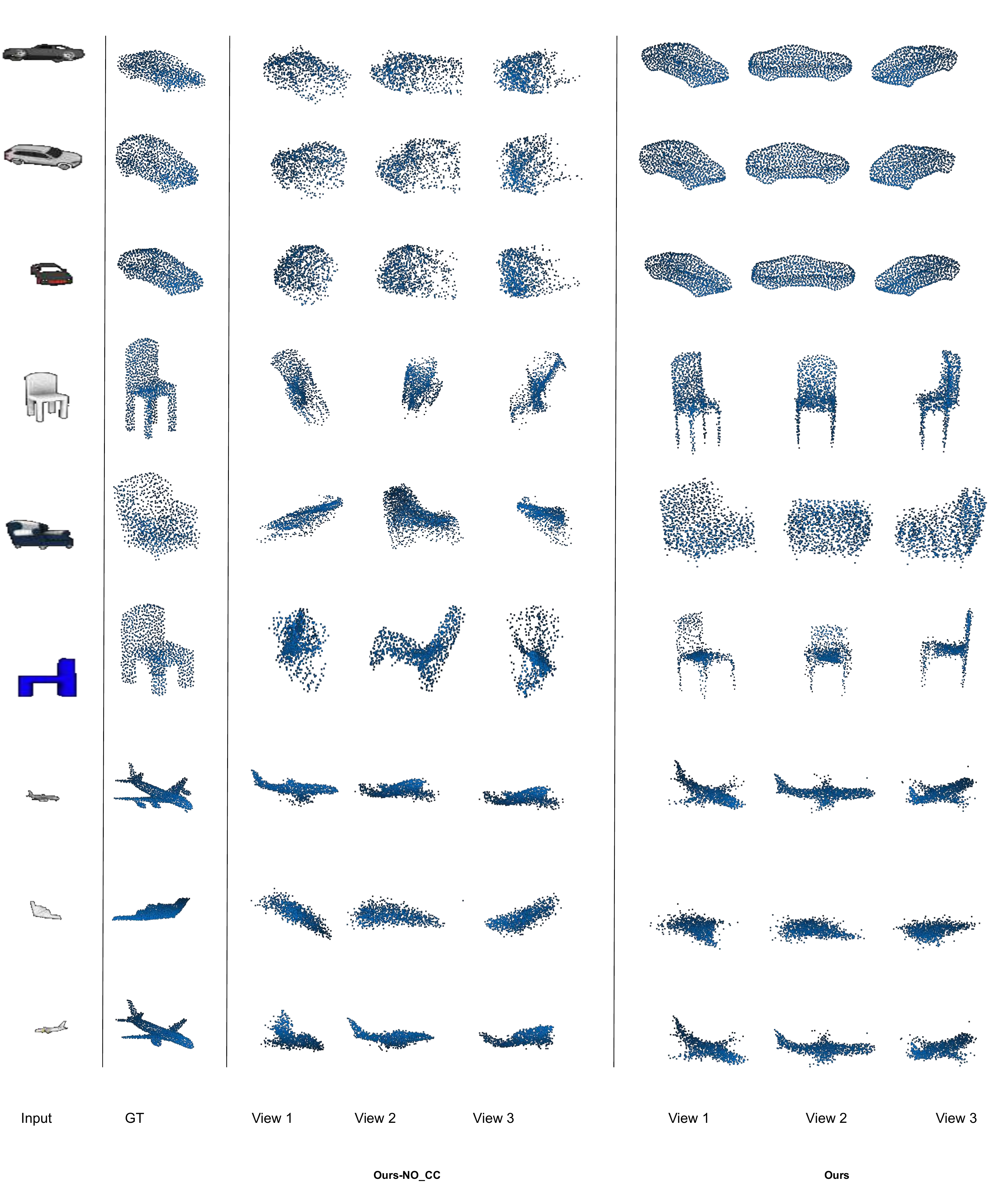}
    \caption{\textbf{Effect of Cycle Consistency Loss.} The network fails to learn meaningful 3D shapes in the absence of the proposed geometric and pose cycle consistency losses. The reconstructions closely match the input when the losses are utilized.}
    \label{fig:consistency_loss}
\end{figure*}

\begin{figure*}
    \centering
        \includegraphics[width=1.0\linewidth]{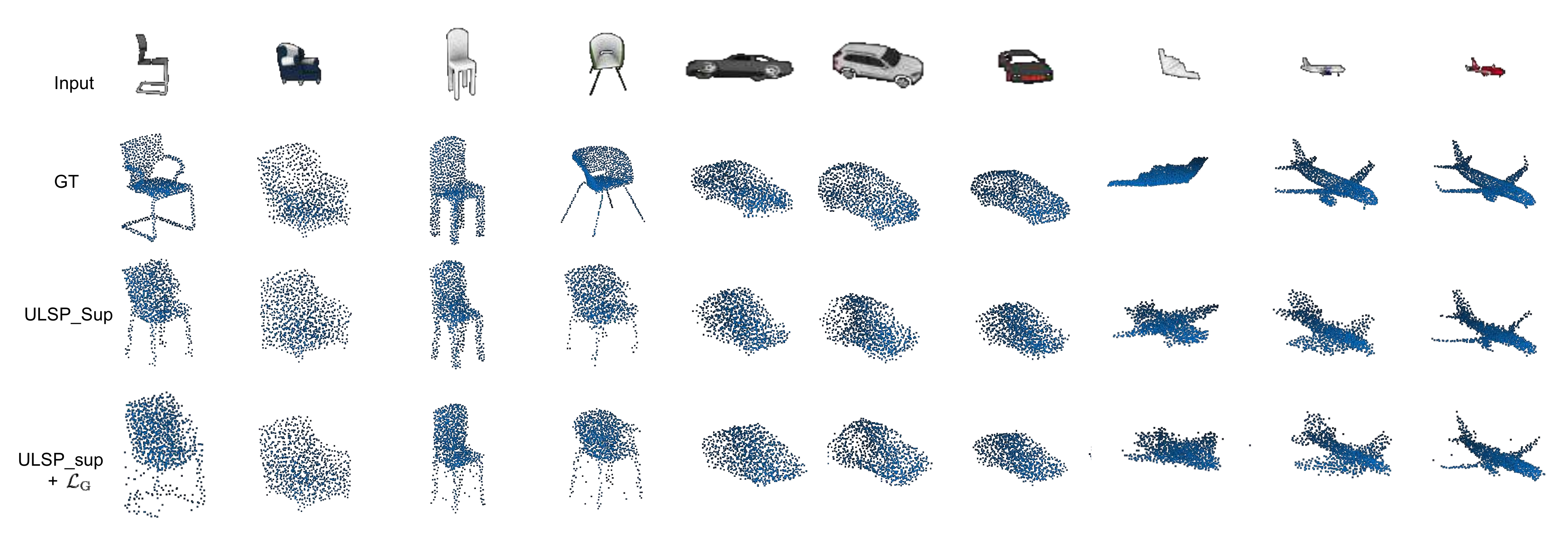}
    \caption{\textbf{Portability of Proposed Loss.} We employ the proposed geometric cycle consistency loss atop the pose-supervised ULSP approach. We observe a significant improvement in the reconstruction quality, suggesting the portable nature of the proposed loss.}
    \label{fig:ULSP_geometric}
\end{figure*}

Fig. 3 and Tables 1 and 3 in the main submission demonstrate the efficacy of the nearest neighbours consistency loss. Here, we analyze the effect of the number of chosen nearest neighbours for each image. Table~\ref{tab:nearest_neighbours_comparison} and Fig.~\ref{fig:Ablation study on Nearest Neighbors} present quantitative and qualitative comparison respectively of reconstruction performance for different number of neighbours. We observe a significant improvement when just a single image is utilized. The performance improves or remains nearly same as more number of images are considered. When more than $3$ images are used in loss calculation, we observe a drop in performance. This behaviour is consistent with our expectations, since the farther nearest neighbours have lower geometric similarity with the input image. 
\begin{table*}
    \centering
    \begin{tabular}{lllllll}
    \toprule
    \multirow{2}{*}{Neighbours} & \multicolumn{2}{c}{Car} & \multicolumn{2}{c}{Chair} & \multicolumn{2}{c}{Aero}\\
    &Chamfer & EMD  & Chamfer & EMD & Chamfer & EMD\\
    \midrule
    0   &6.39   &6.42  &13.58 &16.46 &8.66 & 12.53\\ 
    1   &5.47  &4.93  &10.91 &14.93&8.35 &12.3\\ 
    2   &5.51  &5.29 &10.65 &15.46&7.1 &11.07\\ 
    3   &5.57  &5.16 &10.90 &14.84&8.99 &14.02\\ 
    4   & 5.54  &5.24 &11.93 &16.64&8.65 &13.35\\
    \bottomrule
    \end{tabular}
    \vspace{3mm}
    \caption{\textbf{Effect of Nearest Neighbours.} We examine the effect of number of images of nearest neighbours on the reconstruction metrics. The performance improves or remains nearly same as more number of images are considered. When more than $3$ images are used in loss calculation, we observe a drop in reconstruction performance due to the increased disparity between neighbours and input image.}
    \label{tab:nearest_neighbours_comparison}
\end{table*}


\section{Effect of Symmetry Loss}
Symmetry loss (Section 3.4 of main paper) was proposed as an additional regularization to obtain meaningful 3D reconstructions and to align the reconstructions to a predefined canonical pose. Here, we present quantitative 
results (Table~\ref{tab:sym_loss}) for reconstruction performance with and without symmetry loss. We use both consistency and nearest 
neighbor losses for both the methods. We observe that symmetry loss is crucial in getting reasonable 
reconstructions for the airplane category. It does not affect the performance for the car category while it 
has a negative impact on chair reconstructions. Similar trends were observed on the validation set too. Based on
these observations, we choose the best combination for Ours-NN model. We use the symmetry loss only for the 
airplane category in Ours-NN.
\begin{table}
    \centering
    \scalebox{0.9}{
    \begin{tabular}{lllllll}
        \toprule
        \multirow{2}{*}{Method}   &\multicolumn{3}{c}{Chamfer} &\multicolumn{3}{c}{EMD} \\
         & Car & Chair & Aero & Car & Chair & Aero \\
         \midrule
        Ours-No-Sym & 5.48  & 10.91 & 7.91 & 4.95  & 14.93  & 13.98 \\
        Ours-Sym  & 5.72 & 12.34 & 7.11 & 5.24 & 16.67 & 11.07 \\
        \bottomrule
    \end{tabular}}
    \vspace{3mm}
    \caption{\textbf{Effect of Symmetry Loss.} Symmetry loss is crucial for effective reconstructions on airplane
    category. We choose the best settings from the ablation  for each category in Ours-NN model.}
    \label{tab:sym_loss}
\end{table}

\section{Results on Point Correspondence}
\begin{figure*}
    \centering
    \includegraphics[width=0.6\linewidth]{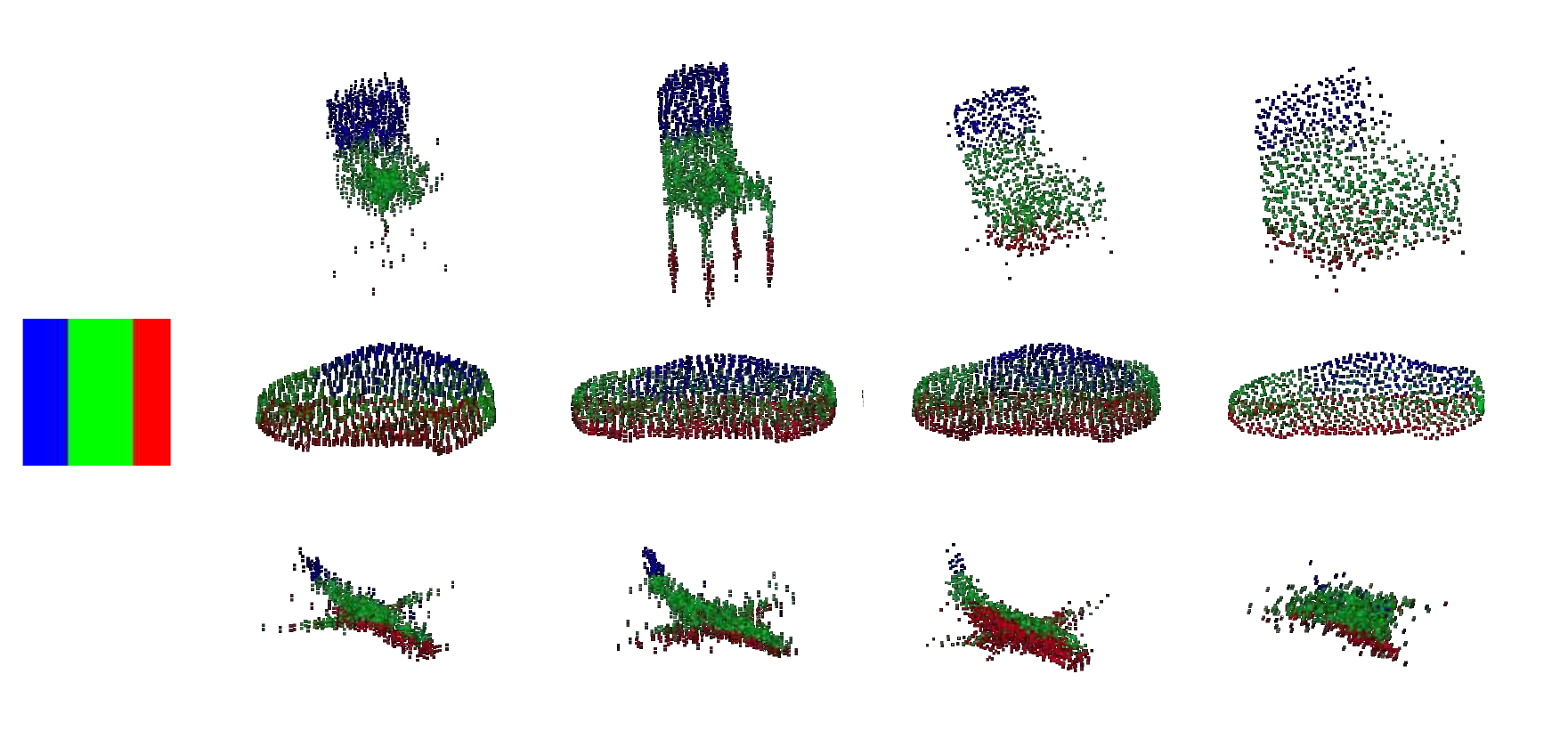}
    \caption{\textbf{Point Correspondence.}  Similar indices in the point cloud are visualized with the same color. The reconstructions exhibit high point correspondence across models.}
    \label{fig:point_corresp}
\end{figure*}

We observe that the reconstructed point clouds have dense point-wise correspondence. That is, points with similar indices in the regressed list of points are present in semantically simillar regions. To visualize this, we use a colored UV map to obtain point correspondences on the point cloud. Fig.~\ref{fig:point_corresp} depicts the UV mapped point clouds. We observe that points with similar color are grouped together and have correspondence across different samples.

\section{Results on Inference Stage Optimization} 
Fig. 4 of the paper demonstrates that ISO results in significant improvement in correspondence of the reconstructions to the input image. We present the corresponding quantitative results in Table~\ref{tab:ISO_metrics}. The metrics are consistent with our observations that the point cloud structure remains intact in occluded regions while closely matching the input image in the visible regions. 

\begin{table}
    \centering
    \scalebox{0.9}{
    \begin{tabular}{llllll}
    \toprule
    \multirow{2}{*}{Categ.}   &\multirow{2}{*}{Method}  &\multirow{2}{*}{Chamfer} &\multirow{2}{*}{EMD} \\
    & & & & \\
    \midrule
    \multirow{2}{*}{Car}     & Ours-NN            & 5.47	& 4.93 \\
                             & Ours-NN post ISO   & 5.49	& 5.01 \\ \midrule
    
    \multirow{2}{*}{Chair}   & Ours-NN            & 10.91	& 14.93	\\
                             & Ours-NN post ISO   & 15.32	& 17.79 \\\midrule
 \multirow{2}{*}{Aero}   & Ours-NN            & 7.1	& 11.07	\\
                             & Ours-NN post ISO   & 7.62	& 11.09 \\
    \bottomrule
    \end{tabular}}
    \vspace{3mm}
    \caption{\textbf{Quantitative Analysis of ISO.} Chamfer and EMD metrics before and after inference stage optimization are comparable. This indicates that the point cloud structures are not degraded in occluded regions due to ISO.} 
    \label{tab:ISO_metrics}
\end{table}

\section{Results on Color Prediction}
Since our networks predict colored point clouds, we present qualitative and quantitative results on it in Fig.~\ref{fig:color_metrics} and Table~\ref{tab:color_metrics}. Due to the absence of good ground-truth for evaluation of color prediction on point clouds, we project our reconstructions from $10$ randomly sampled view-points and perform comparison in the 2D domain. We observe a greater correspondence to the input image in our projections compared to those of the pose supervised DIFFER approach, particularly in the case of car category. Since the color metrics are dependent on the quality of our reconstructions, DIFFER has improved performance in the chair category, while we outperform it in the car category. 

\begin{figure}
    \centering
        \includegraphics[width=1.\linewidth]{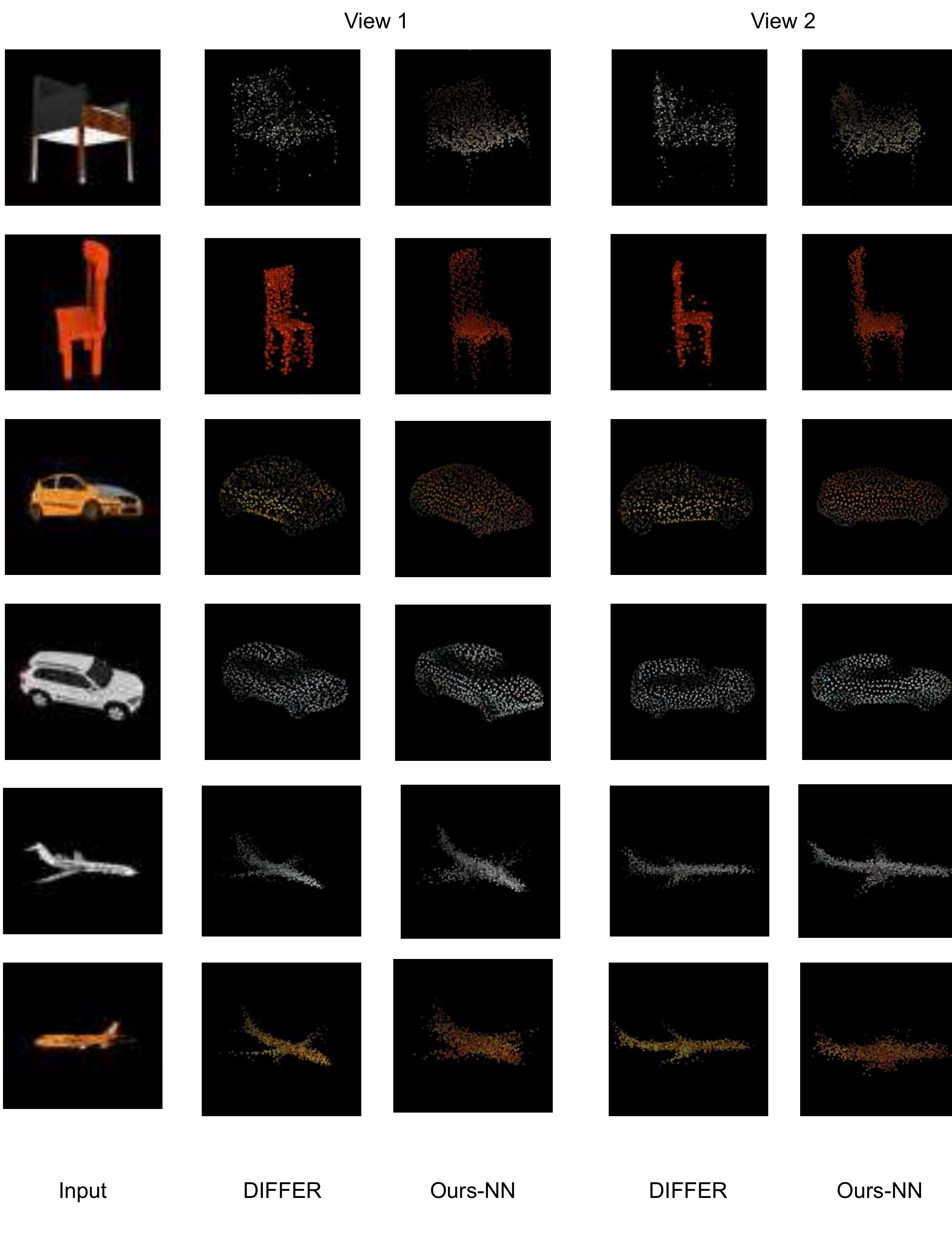}
    \caption{\textbf{Colored Point Cloud Reconstruction.} We compare the colored point cloud reconstructions of DIFFER and our approach. We achieve higher correspondence in color to the input image compared to DIFFER.}
    \label{fig:color_metrics}
\end{figure}

\begin{table}
    \centering
    \begin{tabular}{llll}
        \toprule
        Method      & Car   & Chair & Aero \\
        \midrule
        DIFFER      & 8.59	& 12.81 & 4.69\\
        Ours-CC  & 8.58  & 14.19 & 4.8\\
        Ours-NN     & 8.09	& 13.51 & 4.77 \\
        \bottomrule
    \end{tabular}
    \vspace{3mm}
    \caption{\textbf{Color Metrics.} We present the $\mathcal{L}_2$ distance between predicted projections and ground-truth images to evaluate color prediction. We either outperform or perform comparably to the pose supervised DIFFER approach.}
    \label{tab:color_metrics}
\end{table}

\section{Failure Cases}
\begin{figure}
    \centering
    \includegraphics[width=1.\linewidth]{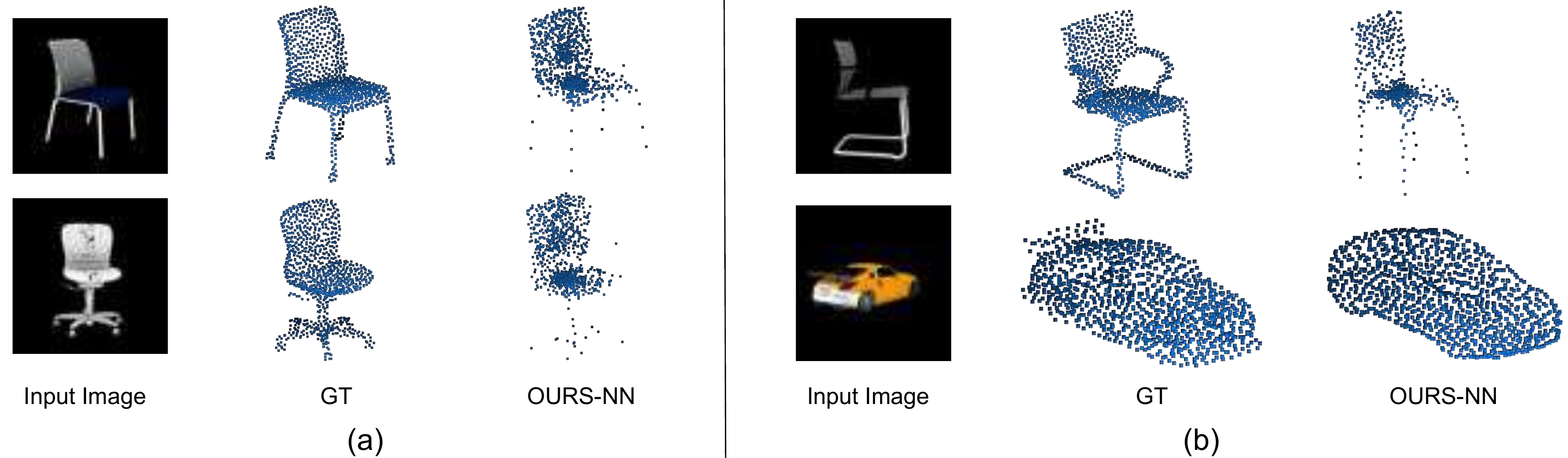}
    \caption{\textbf{Failure cases.} (a) Points are clustered with very few points being used for thin structures like the legs of the chair. 
             (b) Details like car spoilers and complex chair legs/handles are not accurately reconstructed.}
    \label{fig:failure_modes}
\end{figure}

Fig.~\ref{fig:failure_modes} presents a few failure cases.
Some reconstructions
have high density clusters leaving very few points to model the thinner structures 
(Fig.~\ref{fig:failure_modes}(a)). Clusters in airplane category lead to reconstructions with thin structures.
However, we note that such failure modes are also observed in earlier point cloud reconstruction literature~\cite{navaneet2019differ} and addressing these forms an important future work.
Our approach also fails to accurately model certain structures like the spoilers in cars and complex leg and handle structures in chairs (Fig.~\ref{fig:failure_modes}(b)). Training with larger number of such examples might help alleviate the problem.

\section{Network Architecture}

Details of our reconstruction and pose network architectures are provided in Tables~\ref{tab:arch_dual_branch} and ~\ref{tab:arch_pose}. We use a dual branch reconstruction network similar to DIFFER~\cite{navaneet2019differ} for reconstructing point locations and color values. The structure branch of the reconstruction network and the pose network have similar architecture except for the output layer. We use the output of the $D_{s1}$ (Table ~\ref{tab:arch_dual_branch}) layer our reconstruction network as the embedding to obtain the nearest neighbours in our experiments. 

\begin{table}
\centering
\begin{tabular}{|c|c|c|c|c|}
\hline
S.No. & Layer   & \begin{tabular}[c]{@{}c@{}}Filter Size/\\ Stride\end{tabular} & Output Size  \\ \hline\hline
\multicolumn{4}{|c|}{\textbf{Structure Branch}}
\\ \hline
$E_{s1}$     & conv   & 3x3/2                                                         & 32x32x32   \\
$E_{s2}$     & conv   & 3x3/2                                                         & 16x16x64     \\
$E_{s3}$     & conv   & 3x3/2                                                         & 8x8x128    \\
$E_{s4}$     & conv   & 3x3/2                                                         & 4x4x256    \\
$D_{s1}$    & linear & -                                                             & 128          \\
$D_{s2}$     & linear & - & 128         \\
$D_{s3}$     & linear & - & 128         \\
$D_{s4}$     & linear & - & 1024*3      \\ \hline
\multicolumn{4}{|c|}{\textbf{Color Branch}}
\\ \hline
$E_{c1}$     & conv   & 3x3/2                                                         & 32x32x32   \\
$E_{c2}$     & conv   & 3x3/2                                                         & 16x16x64     \\
$D_{c1}$    & linear & -                                                             & 128          \\
$D_{c2}$     & linear & - & 128         \\
$D_{c3}$     & linear & - & 128         \\
$D_{c3}$     & concat($D_{s3},D_{c3}$) & - & 256 \\
$D_{c4}$     & linear & - & 128         \\
$D_{c4}$     & linear & - & 1024*3      \\ \hline
\end{tabular}
\vspace{3mm}
\caption{\textbf{Reconstruction Network Architecture.} We use dual branch network architecture for regressing point locations and color as it is shown to be highly effective~\cite{navaneet2019differ}}
\label{tab:arch_dual_branch}
\end{table}

\begin{table}
\centering
\begin{tabular}{|c|c|c|c|c|}
\hline
S.No. & Layer   & \begin{tabular}[c]{@{}c@{}}Filter Size/\\ Stride\end{tabular} & Output Size  \\ \hline\hline
$E_{s1}$     & conv   & 3x3/2                                                         & 32x32x32   \\
$E_{s2}$     & conv   & 3x3/2                                                         & 16x16x64     \\
$E_{s3}$     & conv   & 3x3/2                                                         & 8x8x128    \\
$E_{s4}$     & conv   & 3x3/2                                                         & 4x4x256    \\
$D_{s1}$    & linear & -                                                             & 128          \\
$D_{s2}$     & linear & - & 128         \\
$D_{s3}$     & linear & - & 128         \\
$D_{s4}$     & linear & - & 2      \\ \hline
\end{tabular}
\vspace{3mm}
\caption{\textbf{Pose Network Architecture.} We use an architecture similar to reconstruction network except for the output layer. In the pose prediction network, two values corresponding to azimuth and elevation parameters of the camera are regressed.}
\label{tab:arch_pose}
\end{table}

\end{document}